# Research on Defect Detection Method of Motor Control Board Based on Image Processing

Jingde Huang[1],*, Zhangyu Huang[2], Chenyu Li[3], Jiantong Liu[1]

[1]School of Intelligent Manufacturing & Aeronautics, Zhuhai College of Science and Technology, Zhuhai 519041, China
[2]Faculty of Innovation Engineering, Macau University of Science and Technology, Macau 999078, China
[3]School of Computer, Zhuhai College of Science and Technology Zhuhai 519041, China

**Abstract**

The motor control board has various defects such as inconsistent color differences, incorrect plug-in positions, solder short circuits, and more. These defects directly affect the performance and stability of the motor control board, thereby having a negative impact on product quality. Therefore, studying the defect detection technology of the motor control board is an important means to improve the quality control level of the motor control board. Firstly, the processing methods of digital images about the motor control board were studied, and the noise suppression methods that affect image feature extraction were analyzed. Secondly, a specific model for defect feature extraction and color difference recognition of the tested motor control board was established, and qualified or defective products were determined based on feature thresholds. Thirdly, the search algorithm for defective images was optimized. Finally, comparative experiments were conducted on the typical motor control board, and the experimental results demonstrate that the accuracy of the motor control board defect detection model-based on image processing established in this paper reached over 99%. It is suitable for timely image processing of large quantities of motor control boards on the production line, and achieved efficient defect detection. The defect detection method can not only be used for online detection of the motor control board defects, but also provide solutions for the integrated circuit board defect processing for the industry.

**Keywords:** Motor control board, Defect detection, Defect feature, Feature recognition, Image Processing.

## 1. Introduction

The motor is the executing mechanism and key component of automation equipment. The normal operation of the motor requires various types of motor drive control boards. Essentially, the motor control board is also a type of circuit board that can play a control role. The motor control board defect detection refers to the inspection of the function, performance, and appearance quality of motor control boards already soldered with components, such as inconsistent color differences, incorrect plug-in positions, and solder short circuits. If these defects are not detected and repaired in time, they will severely affect the motor control board's performance and stability, negatively impacting product quality. The detection technology of the motor control board finished products is an important link in ensuring product quality. Currently, commonly used methods for detecting finished motor control boards include appearance inspection, signal injection, substitution, non-online measurement, parameter testing, waveform investigation, intuitive viewing, signal tracing, communication short circuit testing, and separation testing [1-3]. Among the above detection methods, the appearance inspection is a simple and effective method, which mainly detects whether there are cracks, short circuits, open circuits, exposed copper wires, and other defects on the motor control boards through observation and comparison. Appearance inspection can be divided into two methods: manual appearance inspection and automatic appearance inspection. Artificial appearance inspection refers to the use of tools such as the naked eye or magnifying glass by the operator to inspect the motor control boards one by one and determine whether they are qualified based on standards or experience [4-6]. Artificial appearance inspection has the advantages of simple operation and low equipment investment, but it also has the disadvantages of strong subjectivity, low efficiency, and a high rate of misjudgment. Automatic appearance detection refers to the use of machine vision technology and image processing technology to quickly and accurately identify and analyze the motor control boards, and compare them with reference templates to identify defects and generate reports. Automatic appearance inspection has the advantages of strong objectivity, high efficiency, and high accuracy, but it also has disadvantages such as high equipment costs and complex maintenance.

*Corresponding author.
*Email addresses*: jdh925@zcst.edu.cn (Jingde Huang).

With the development of machine vision detection technology, an increasing number of high-speed 3D visual scanning systems and thermal imaging systems are widely used in various fields such as manufacturing, military, medical. For the application of image processing in defect detection, the YOLO series algorithms have played an important role in industrial defect detection [7]. They transform visual defect detection into a regression problem for processing, reflecting the powerful technical advantages of deep learning and receiving much attention and welcome in the field of computer vision. The adaptive multi-scale YOLO algorithm has demonstrated great advantages in the field of image processing. Compared with traditional multi-stage processing methods, this algorithm simplifies the overall workflow and achieves the best balance between processing accuracy and speed [8]. To improve the performance of the framework, Parupalli S [9], Subramani S [10] and others have improved YOLOv2 YOLOv3, By improving the network structure, multi-scale prediction has been achieved. The currently stable YOLOv7 [11] adopts an efficient long-range network (ELAN) module as the feature extraction unit, which improves processing efficiency and accuracy. However, in practical engineering applications, there are still some challenges in machine vision-based defect detection technology [12-14]. For example, some subtle defects may either go undetected or be difficult to accurately identify in complex backgrounds. At the same time, the traditional motor control board finished product inspection methods not only have low efficiency in situations where the motor control board production is large and assembly line speed is fast, but also easily lead to visual fatigue and errors, affecting production efficiency. Therefore, improving the defect detection accuracy of machine vision has always been a key research focus in the field of motor control board defect detection [15-17]. Currently, domestic and international scholars have conducted extensive theoretical and experimental explorations on the application of machine vision technology in defect detection, and have proposed numerous valuable detection methods. The detection technology based on artificial intelligence has developed rapidly in recent years. In 2017, Liu Tao and Yin Shibin et al.[18] designed a fast visual detection system primarily used for detecting the width and length of items. In 2023, Hu Xinyu and Kong Defeng et al.[19] developed a fast-moving small-targets detector with high accuracy and speed. In 2025, Whelan M and Park Y [20] created a visual-based 3D measurement system for measuring and analyzing product contours and shapes.

In summary, the defect detection method for control boards based on machine vision has a certain foundation, but there is still a need for confirmatory testing in the rapid detection and recognition of minor defects. Moreover, many researchers focus on the algorithm field, and there is a lack of specific exploration in engineering applications such as quality inspection of production line products. In engineering practice, the variation of grayscale values has a significant impact on the processing effect of dark details on the control board. At the same time, the color consistency and quality of the control board are crucial for ensuring the reliability and performance of electronic components, so color perception is also an important foundation of image processing technology. This paper focuses on the problems in defect detection of the motor control board mentioned above. Its highlights lie in the application of image processing technology in defect detection. Firstly, a color difference recognition method for motor control boards was proposed. Secondly, a defect feature extraction and recognition method for motor control boards was proposed. Thirdly, the search algorithm for defective images was optimized. Finally, the proposed model and method were validated through experiments. The proposed research method is of great significance for continuously improving product quality and production efficiency, and promoting the safe and efficient development of production.

## 2. Related work

### 2.1. Binocular calibration method for image acquisition

Camera calibration is a crucial step in extracting three-dimensional spatial information from two-dimensional images and is also the foundation for achieving machine vision system calibration [21-22]. Firstly, the internal and external parameter matrices of the camera are obtained through the binocular calibration method, and then the distortion of the image is corrected to obtain an undistorted digital image, providing the technical basis for the machine vision system to acquire accurate image parameters. The camera parameters describe the geometric relationship between the image position and the corresponding points on the surface of the spatial object, and the transformation relationship between the camera coordinates and the image coordinates is from three-dimensional to two-dimensional. The coordinate relationship of the camera imaging model is shown in Fig. 1. The intersection point between the line *OP* connecting the optical center *O* and any point *P* in the world coordinate system and the image plane is the projection position of any point *P* on the image, where $O_w-$ *represents* the world coordinate system, $O_c-X_cY_cZ_c$ represents the camera coordinate system, $O-xy$ represents the image coordinate system, $uv$ represents the pixel coordinate system, *p(x,y)* represents the imaging point of point *p* in the image, and *f* represents the camera focal length.

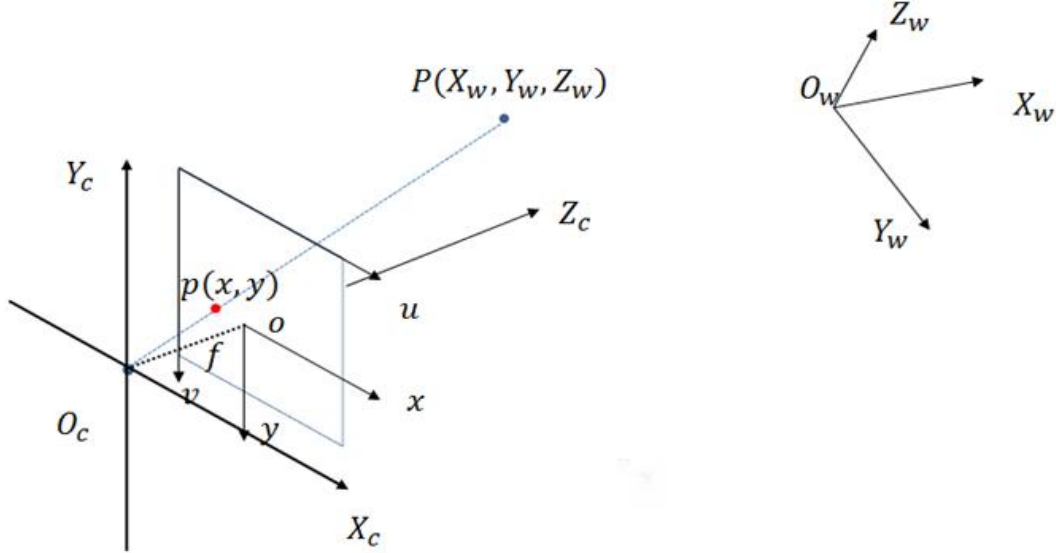

**Fig. 1.** Coordinate relationship of camera imaging model.

In order to obtain the target image in all directions and preserve the subtle features of the image, this paper uses a binocular vision system to obtain the motor control board detection image. Therefore, it is necessary to confirm the relative position relationship between the two cameras in the binocular vision system and calibrate the relative relationship between the coordinate systems of the left and right cameras. Binocular calibration is mainly used to determine the correspondence between the image coordinates of all coordinate points and the world coordinates, and to obtain the correct three-dimensional pose information of the target image. Firstly, use left and right cameras to capture images of the same calibration board in different poses, forming 15 stereo images; Secondly, initialize the calibration board description file and camera parameters, and save the initial pose of the left and right calibration boards as well as an array of calibration point image coordinates; Thirdly, obtain the initial pose of the calibration board relative to the left and right cameras and the image coordinates of its calibration points, and save them separately in the initialized array; Finally, the initial calibration of both eyes is completed through the calibration algorithm. The initial calibration results are shown in Table 1, $S_x$ represents the width of a single pixel, $S_y$ represents the height of a single pixel, $(u_0,v_0)$ is the midpoint of the imaging plane, and *Err* represents the average error.

Table 1. Calibration parameters for binocular cameras

| Cameras Parameters | $S_x$ (μm) | $S_y$ (μm) | f (mm) | $u_0$ | $v_0$ | Err |
|---|---|---|---|---|---|---|
| Left camera | 2199.9132 | 2199.9985 | 0 | 662.418 | 453.05 | 0.247863 |
| Right camera | 2213.9585 | 2214.0021 | 0 | 645.432 | 467.28 | 0.247863 |

## 2.2. Optimization of search algorithm for images with defects

In engineering, if defects in the motor control board cannot be detected and repaired in a timely manner, it will seriously affect the performance and stability of the equipment. Therefore, it is necessary to quickly, effectively, and accurately detect these defects [23]. Considering factors such as high output of motor control boards and fast assembly line speed, this article conducted in-depth research on the screening of defect images. The motor control board is generally printed with a large amount of redundant information such as numbers, patterns, and borders. In order to compress the image data and improve the efficiency and accuracy of defect pattern recognition algorithms, this paper uses Ant colony optimization (ACO) algorithm to optimize and segment the regional features in the original image, ensuring the edge characteristics of the image and achieving high segmentation accuracy [24-25]. The optimization process of ACO is shown in Fig. 2. The ACO algorithm can accelerate the search speed of defect images based on the historical images of the motor control board. By using the boundary, it can ensure that the model parameters do not exceed the set range, avoid unreasonable search results, and help the defect recognition model better adapt to different data distributions and changes.

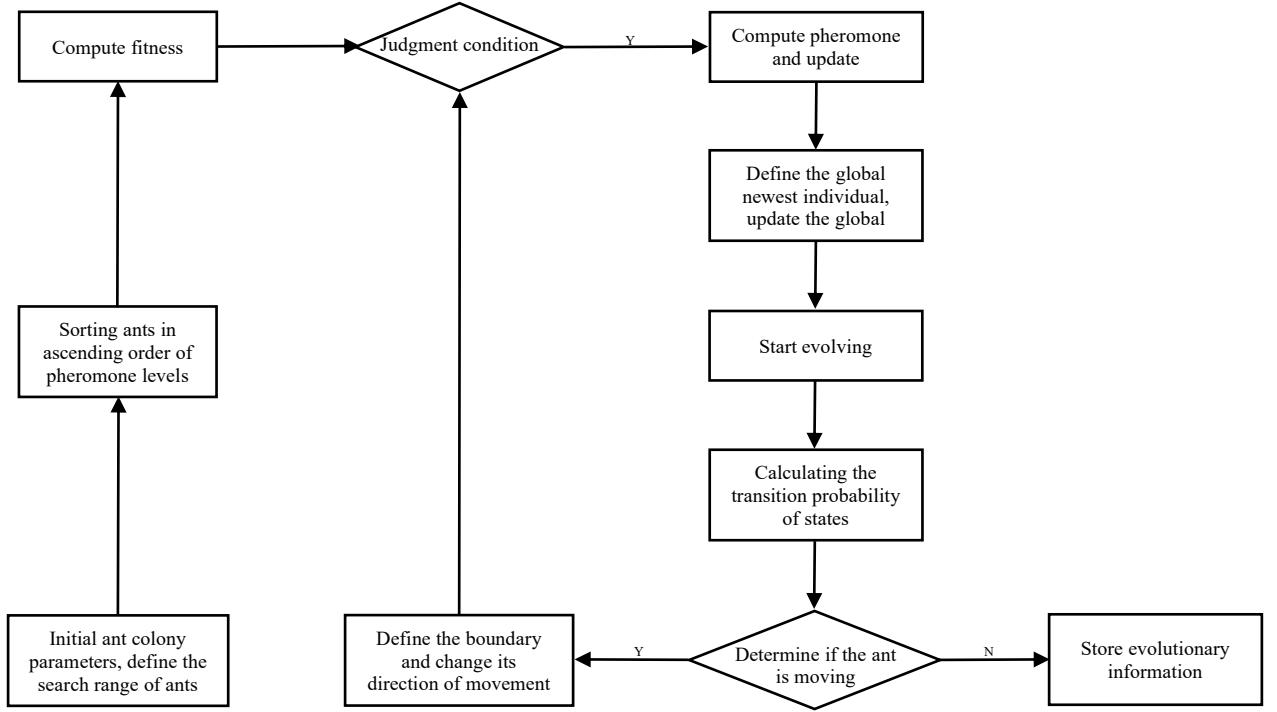

**Fig. 2.** The optimization process of ACO.

The ACO algorithm records all the city numbers passed by $g$ in search order $R^g$. Assuming that the ant is currently in city $x$, the probability of searching $P_g(x, y)$ for the next city $y$ is as follows:

$$P_g(x, y) = q \begin{cases} \dfrac{[\tau(x,y)]^\alpha [\eta(x,y)]^\beta}{\sum_{u \in J_g^{(x)}} [\tau(x,u)]^\alpha [\eta(x,u)]^\beta} & y \in J_g(x) \\ 0 & otherwise \end{cases} \quad (1)$$

Where, $P_g(x, y)$ is the transition probability, $\tau(x, y)$ is the concentration of pheromones between city $x$ and city $y$, $\eta(x, y)$ is the path-length from city $x$ to city $y$, $J_g^{(x)}$ is the unexplored city of the $x$ ant in the $g$ population, and the number of ants $\beta$ and $\alpha$ are control parameters.

Updating pheromone trajectories is the best way to improve solutions. Path updates include local updates and global updates, with the local update formula as follows:

$$\tau(x, y) = (1 - \rho) \cdot \tau(x \cdot y) + \sum_{g=1}^{m} \Delta \tau_g(x, y) \quad (2)$$

Where $\rho$ $(0 < \rho < 1)$ is the evaporation rate of pheromones, and $\Delta \tau_g(x, y)$ is the amount of edge pheromones added between city $x$ and city $y$.

$$\Delta \tau_k(i, j) = \begin{cases} H \times (C_k)^{-1} & (x, y) \in R^g \\ 0 & otherwise \end{cases} \quad (3)$$

Where, $H$ is the total amount of pheromones, and $C_k$ is the distance between the ant colony and $\pi_g$ in $\Delta t$.

## 2.3. Defect detection system based on intelligent vision

At present, defect detection technology based on machine vision has moved from theoretical research to practical application. It is necessary to quickly and effectively detect the surface of motor control boards. This paper focuses on exploring the application of machine vision technology in defect detection and applying it to the actual production of motor control boards, improving product quality and production efficiency. In the production or assembly process of motor control boards or integrated circuit boards, traditional manual inspection methods are not only inefficient, but also often result in missed or false detections, making it difficult to meet the current demand for high efficiency in industrial production. The integration of computer network technology with industrial production has led to the development of numerous intelligent detection systems. Based on this, a practical visual inspection system for motor control board defects is proposed. The system uses a Programmable Logic Controller (PLC) as the control core and combines machine vision technology to identify defect information on the motor control board, achieving automated and intelligent quality inspection results.

In engineering practice, PLC is the main device for industrial automation, with complete input/output functional modules and flexible configuration, which can be used to drive the robotic arm of intelligent visual inspection system. The intelligent visual inspection system established in this paper is mainly controlled by PLC, including transmission module, visual inspection and recognition module, intelligent robotic arm module, human-machine interface and database, etc. Its workflow is illustrated in Fig. 3.

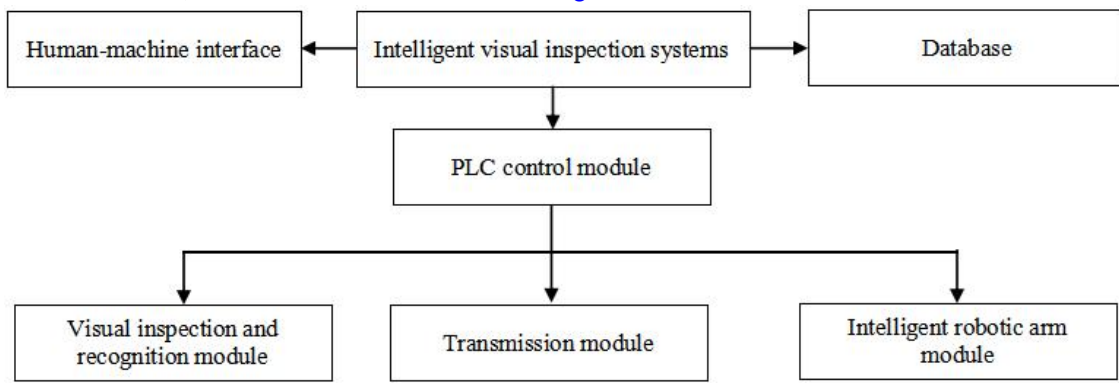

**Fig. 3.** Intelligent visual inspection system workflow.

Firstly, the operator deploys detection tasks to the intelligent visual inspection system through a human-computer interaction interface; Secondly, when the PLC receives the information, it sends control instructions to control the operation of the transmission module, driving the control board to move forward in a pipeline form; Thirdly, when the visual recognition module captures the control board image, it enhances the image, extracts the defect area for analysis and recognition, obtains defect information, and performs defect feature recognition; Finally, the intelligent robotic arm module is used to eliminate non-conforming products.

## 3. Methods

### 3.1. Noise processing of digital image

During the process of digital image acquisition, conversion, transmission, and storage, image quality may decrease due to various factors, which can affect the detection accuracy of machine vision. In the process of image processing for the parts being tested, noise suppression becomes crucial. The purpose of noise suppression is to reduce the grayscale information of images generated by non-object sources, improve image quality, and achieve accurate detection [26-29]. According to the different requirements of image processing, different denoising methods need to be selected in order to achieve satisfactory results. This is because each filter has its advantages and disadvantages. Firstly, for mean filtering, considering that most of the noise in the image is random and uncertain, usually manifested as a sudden change in grayscale, using mean filtering can effectively weaken the noise. Moreover, as the neighborhood radius increases, the smoothing effect becomes stronger. However, at the same time, due to the sudden changes in grayscale values at the edges and details of the areas of interest to us, an excessively large mean template can cause blur of the edges. Secondly, median filtering selects the median by sorting the grayscale values in the mask, and replaces the central pixel values with the median. Therefore, median filtering can effectively maintain the slope function and the step function unchanged, so median filtering has a good effect for extracting a certain uniform distribution of noise and salt and pepper noise. The overall grayscale value of the digital image tends to a fixed value, making the resulting image smoother. Thirdly, for Gaussian filtering, in image Gaussian smoothing, the central pixel has greater weight than the surrounding pixels in the template, because pixels near the center pixel have higher importance. The function of Gaussian filtering describes the probability density of normally distributed random variables, so it is very effective in suppressing normally distributed noise. After comparison, it was found that Gaussian filtering preserves edge information completely, and the actual display effect is better than median filtering and mean filtering. In terms of time consumption, mean filtering and median filtering have the shortest time consumption. However, considering that mean filtering and median filtering blur the edges to a certain extent, in order to ensure the accuracy of detection, this paper chooses Gaussian filtering as the denoising method for the motor control board detection.

This paper uses Gaussian filtering to weighted average the entire image. The weight value of Gaussian filtering is obtained by the following formula.

The one-dimensional Gaussian distribution:

$$G(x) = \frac{1}{\sqrt{2\pi}\sigma} e^{-\frac{x^2}{2\sigma^2}} \tag{4}$$

The two-dimensional Gaussian distribution:

$$G(x,y) = \frac{1}{\sqrt{2\pi}\sigma} e^{-\frac{x^2+y^2}{2\sigma^2}} \tag{5}$$

Where, $\sigma$ is the Gaussian coefficient; $x$ is the row-coordinates, and y is the column-coordinates.

In Gaussian filtering, pixels closer to the center pixel have higher importance, so in the process of weighted averaging, pixels closer to the center will have greater weight than other pixels. The Gaussian filtering template can be represented by a matrix, as shown in Fig. 4.

| 1/16 | 2/16 | 1/16 |
|------|------|------|
| 2/16 | 4/16 | 2/16 |
| 1/16 | 2/16 | 1/16 |

**Fig. 4.** Gaussian filter template

According to the defect type of the motor control board, different feature recognition operators can be used. Considering the application characteristics of specific operators, in order to achieve the best recognition effect, this article intends to select and adjust corresponding operators according to specific application scenarios and task requirements. Taking the template matching operator as an example, suppose there is an image $I$ and a template $T$, with an image size of $M*N$ and a template size of $m*n$. Given a subregion $I(x,y)$ of an image, whose size is the same as template $T$, i.e. $I(x,y)$ is a submatrix $m*n$ starting from position $(x,y)$ in image I. The calculation of correlation coefficient is shown in formula (6):

$$R(x,y) = \frac{\sum_{i=0}^{m-1}\sum_{j=0}^{n-1}\left[I(x+i,y+j)-\bar{I}xy\right]\left[T(i,j)-\bar{T}\right]}{\sqrt{\sum_{i=0}^{m-1}\sum_{j=0}^{n-1}\left[I(x+i,y+j)-\bar{I}xy\right]^2}\sqrt{\sum_{i=0}^{m-1}\sum_{j=0}^{n-1}\left[T(i,j)-\bar{T}\right]^2}} \tag{6}$$

Where, $\bar{I}xy$ represents the average value of the sub-region $I(x,y)$, calculated using formula (7):

$$\bar{I}(x,y) = \frac{1}{mn}\sum_{i=0}^{m-1}\sum_{j=0}^{n-1} I(x+i,y+j) \tag{7}$$

Where, $\bar{T}$ represents the average value of template $T$, and the calculation is shown in formula (8):

$$\bar{T} = \frac{1}{mn}\sum_{i=0}^{m-1}\sum_{j=0}^{n-1} T(i,j) \tag{8}$$

By calculating the value of $R(x,y)$ in the entire image, the position corresponding to the maximum value is the best matching position. This value represents the similarity between the template and the graphic subregion, with higher values indicating higher similarity.

## 3.2. Gray level changes in digital image

The grayscale change of a visual image refers to the method of changing the grayscale value of the original image one by one pixel according to a certain transformation relationship based on a certain target condition, which can also be understood as the contrast change of the visual image [30-33]. The grayscale changes of visual images can increase the contrast, and make the image more obvious and clear, making it an important means of image enhancement. The grayscale changes of visual images can be divided into linear grayscale changes and nonlinear grayscale changes.

The linear change in grayscale refers to the change in grayscale of a visual image according to a certain linear relationship. The image function is $f(x,y)$ with a grayscale value range of $[a,b]$, and the transformed image function is $g(x,y)$ with a grayscale value range of $[c,d]$. The variation formula is denoted as follows:

$$g(x,y) = k[f(x,y) - a] + c \qquad (9)$$

Where, $k = \dfrac{d-c}{b-a}$ is the slope of the straight line, which represents linear expansion or compression of the input image grayscale.

Linear variation of grayscale to some extent solves the problem of overall image contrast. Although the transformation is uniform, this method is not easy to achieve non-uniform transformation. If the processing method is divided into multiple segments, once the number of segments is too large, it will be quite troublesome to handle. Therefore, it is necessary to introduce a method of nonlinear transformation, which maps functions across the entire range of grayscale values. Generally speaking, a unified transformation function is used for nonlinear transformation to achieve compression and expansion of different grayscale values. The two commonly used nonlinear transformations are logarithmic transformation and exponential transformation.

Logarithmic transformation maps the grayscale range of an image to a logarithmic function. The grayscale of the image can be changed through the properties of the logarithmic function. The logarithmic function can compress the larger range of grayscale values and also expand the smaller range of grayscale values. The function of logarithmic transformation is denoted as follows:

$$g(x,y) = a + \frac{\log[f(x,y) + c]}{\log b} \qquad (10)$$

Where, $a$, $b$ and $c$ are parameters that facilitate adjusting the position and shape of the image. $a$ controls the upper and lower positions of the image, $b$ controls the transformation trend of the image, and $c$ controls the left and right positions.

Exponential transformation maps the grayscale range of an image to an exponential function, which can selectively reduce the contrast of high grayscale areas or enhance the contrast of low grayscale areas based on different parameter values. The function expression for exponential transformation is denoted as follows:

$$g(x,y) = a[f(x,y) + b]^c \qquad (11)$$

Where, $a$, $b$ and $c$ represent the parameters set to adjust the position and the shape of the image.

## 3.3. Grayscale conversion of digital image

The grayscale conversion of digital images is the process of converting color or monochrome images into grayscale images, where the grayscale value of each pixel on the digital image represents the brightness of that pixel. In the grayscale conversion process, monochrome images can be directly converted into grayscale images, while color images require weighted averaging of the pixel values of the three RGB channels to obtain the grayscale values of each pixel. For the RGB image, each pixel is composed of values from three channels: red, green and blue. In order to convert the RGB image into a grayscale image, it is necessary to weighted sum the RGB channel values of each pixel to obtain the corresponding grayscale value.

The processing of the color image uses the formula (12):

$$GrayValue = 0.299 * R + 0.587 * G + 0.114 * B \tag{12}$$

Where $GrayValue$ represents the corresponding grayscale value, $R$, $G$ and $B$ are the pixel values of the three color channels in the RGB image, and the coefficients 0.299, 0.587 and 0.114 correspond to the weighted values of the red channel, green channel and blue channel respectively.

Process the color image of a certain motor control board, and the processed image is shown in Fig. 5.

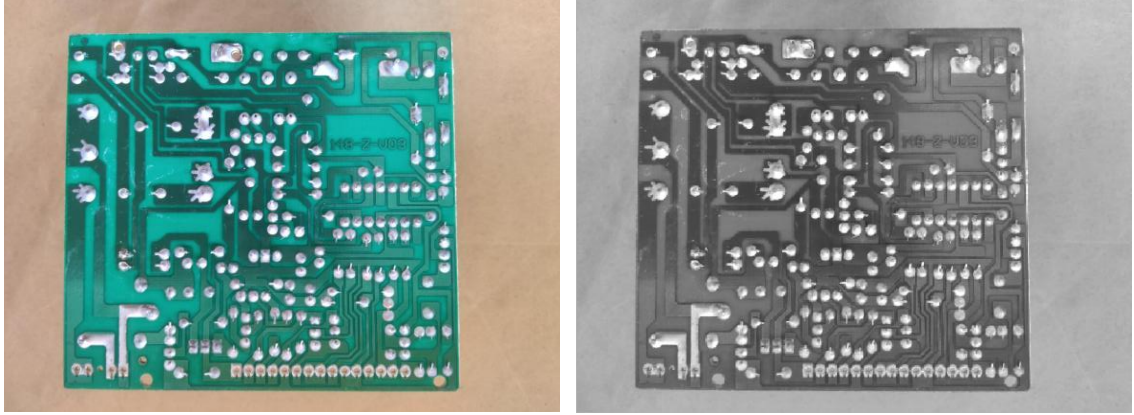

(a) Image before processing          (b) Processed image
**Fig. 5.** Color difference comparison effect of motor control board.

## 3.4. Color processing model and algorithm

The RGB space is composed of R (red ), G (green), and B (blue). The HSV space describes color changes through $H$ (hue), $S$ (saturation), and $V$ (brightness). Considering that the three color components in the RGB space are not directly related to the final color, while the HSV color space can describe colors through three dimensions: color, depth, and brightness, it not only conforms to the way humans perceive colors, but also provides a concise and complete description of colors. Therefore, HSV space is usually used to detect color information in engineering practice [34-36]. The HSV model is typically used for tasks such as color filtering, color segmentation, and object recognition in image processing. By selecting a specific color range in the HSV space, pixels of a specific color can be separated from the image for subsequent processing and analysis.

Before converting the RGB space to the HSV space, the values of R (red ), G (green) and B (blue) need to be divided by 255 to change their range from 0 to 255 to 0 to 1. To effectively detect color differences on the motor control board, this paper decomposes an RGB color image into three grayscale images, representing the brightness information of the R channel, G channel, and B channel, respectively. Firstly, the RGB image is represented as three independent grayscale images, and the pixel values in each grayscale image represent the brightness intensity of the corresponding channel. Secondly, the decompose3 operator is used to process and analyze specific channels in color images. If a motor control board image is captured under red light illumination, the red channel can be used to enhance the contrast and details of the image. If a motor control board image is captured under green light illumination, the green channel can be used to enhance the contrast and details of the digital image [15].

For the average and standard deviation of the grayscale values of the input image in a given region, if $R$ is a region, $P$ is a pixel in $R$, and its grayscale value is $g(p)$, and $F$ represents plane $(F = |R|)$, then the equation set of characteristic values can be defined as follows:

$$Mean := \frac{\sum_{p \in R} g(p)}{F}$$

$$Deviation := \sqrt{\frac{\sum_{p \in R} (g(p) - Mean)^2}{F}} \tag{13}$$

## 3.5. Edge extraction and feature recognition

(1) Edge extraction method

The geometric size detection of the motor control board is crucial for extracting edge features. The objects in the image contain a lot of basic information, such as their shape and size. Object edge detection is performed by finding and filtering points in the image that have significant changes in grayscale values, and converting these points into the edges of the object. Typical edge detection methods include Prewitt gradient method, Sobel gradient method, Roberts gradient method, and Canny detection method [15]. The characteristic of the Prewitt operator is that it does not take the center point as the center of gravity, which makes it more sensitive to the direction of the edges. While suppressing noise appropriately, it also plays a smoothing role on the edges. The Sobel operator is the same as the Prewitt operator, but the difference is that the Sobel operator weights the position influence of pixels. Compared to Prewitt, Sobel effectively reduces the degree of edge blur [37-39]. Considering the fast computational speed of Roberts gradient method, it can better meet the real-time requirements of defect detection in motor control boards. Meanwhile, the appearance shape of the control board has good regularity, and the Roberts gradient method can effectively detect the diagonal edges of the control board. The Roberts gradient method can be represented by formula (14) and (15):

$$G[f(x,y)] = \sqrt{[f(m,n) - f(m+1,n+1)]^2 + [f(m+1,n) - f(m,n+1)]^2} \tag{14}$$

$$\begin{array}{ccc} f(m,n) & \searrow\nearrow & f(m,n+1) \\ & \nearrow\searrow & \\ f(m+1,n) & & f(m+1,n+1) \end{array} \tag{15}$$

From the above formulas, it can be seen that the Roberts operator is not a horizontal or vertical gradient method, but a cross difference calculation method, so it has precise edge positions. However, since the image has not been smooth, it cannot suppress noise. The Roberts operator is suitable for image segmentation with obvious sharp edges and low noise.

The characteristic of the Canny operator is that it adopts dual threshold and retains the maximum central value along the edge direction, otherwise the central point is set to 0. Double threshold refers to setting an upper and lower threshold. If the pixel of an image exceeds the upper threshold, it is considered a boundary, also known as a strong boundary. If the threshold does not exceed the lower threshold, it is considered not a boundary. When the grayscale value of the pixel is between the upper and lower thresholds, the next step is required, called a weak boundary. In engineering practice, the Canny operator effectively suppresses the edges disguised by noise, and refines the edges. This is undoubtedly the best choice for detecting the geometric dimensions and other features of the motor control board using Canny's edge extraction method. The processing process is:

**Step 1:** Calculate the gradient amplitude and direction

Gradient reflects the changes in image pixels. If $G_x$ and $G_y$ are the grayscale values of the digital image in the $x$ and $y$ directions, and $f$ is the original image, the gradient formula is as follows:

$$G_x = \begin{bmatrix} -1 & 0 & 1 \\ -2 & 0 & 2 \\ -1 & 0 & +1 \end{bmatrix} \bullet f \tag{16}$$

$$G_y = \begin{bmatrix} 1 & 2 & 1 \\ 0 & 0 & 0 \\ -1 & -2 & -1 \end{bmatrix} \bullet f \tag{17}$$

The gradient amplitude $M(x,y)$ is:

$$M(x,y) = \sqrt{G_x^2 + G_y^2} \tag{18}$$

The gradient direction $\theta$ is:

$$\theta = \arctan\left(\frac{G_y}{G_x}\right) \tag{19}$$

**Step 2:** Eliminate the spurious response caused by edge detection

After obtaining the gradient amplitude $M(x,y)$, in order to ensure accurate positioning, it is necessary to use the non-maximum suppression method to refine it. The non-maximum suppression method suppresses pixels with non-maximum gradients in the neighborhood. For any pixel, compare its gradient value with the gradient value in the neighborhood. If it is not the maximum gradient value, set the gray level of the pixel to 0. If it is the maximum gradient value, the new suppressed image is represented as follows:

$$N(x,y) = M(x,y) \tag{20}$$

To further reduce erroneous edge points, eliminate weak edges, and retain strong edges, appropriate high threshold $T_H$ and low threshold $T_L$ are selected. If $N(x,y) > T_H$, this point is an edge point, if $N(x,y) < T_L$, it is not an edge point, and if $T_L < N(x,y) < T_H$, it is necessary to consider whether the neighborhood of this point has pixels greater than $T_H$. If there are related pixels, then this point is an edge point, and if there are no related pixels, then this point is not an edge point.

Considering that using the Canny operator to extract edges from the original image is easily affected by some line feature noise, in the case where the lines in the barcode area are all straight lines, if the noisy lines are curved, they cannot be fitted. In the process of fitting a straight line, outliers occupy a significant weight in the least squares method, resulting in inaccurate straight lines. Therefore, this paper uses the Tukey weight function $\omega(\delta)$ to eliminate the influence of outliers as follows:

$$\omega(\delta) = \begin{cases} \left[1 - \left(\frac{\delta}{\tau}\right)^2\right]^2 & |\delta| \leq \tau \\ \frac{\tau}{|\delta|} & |\delta| > \tau \end{cases} \tag{21}$$

Where, parameter $\tau$ represents the distance threshold, parameter $\delta$ represents the distance from point to line. According to formula (21), the farther the distance from point to line, the smaller the weight it occupies.

For the barcode recognition of the motor control board, in order to ensure that the extracted area is a barcode area and correctly identify the category of the control board, this paper adopts a color based area determination method. The barcode is black-white, and the proportion of the white area to the black area is roughly the same. If the value of the black-white area is between 0.7 and 1.5, it is determined as a bar code area, completing the final positioning segmentation [40-42].

(2) Feature recognition method

For image feature recognition of integrated electronic modules under different imaging conditions and at different times, SAD (Sum of absolute differences) is generally used in engineering to calculate the absolute sum of pixel grayscale differences between the template image and the detection image, as shown in formula (22):

$$SAD(x,y) = \frac{1}{n} \sum_{(u,v) \in T} |t(u,v) - f(x+u, y+v)| \tag{22}$$

The above method can only achieve good matching results when the lighting remains constant. Once the lighting intensity changes, the calculated difference will undergo significant changes. Therefore, calculating similarity solely based on point-to-point differences is easily affected by noise. In order to improve the practicality of the method and accurately identify image features, this paper introduces the SSD (Sum of Squared Differences), as shown in formula (23):

$$SSD(x,y) = \frac{1}{n} \sum_{(u,v) \in T} (t(u,v) - f(x+u, y+v))^2 \tag{23}$$

## 3.6. Barcode processing algorithm

To ensure recognition accuracy, when the barcode image on the motor control board is blurry, image enhancement algorithms must be used for optimization. Due to the fact that barcodes consist of many black bars and white spaces, the use of linear grayscale transformation can make the black areas of the image darker and the bright areas brighter, effectively solving the problem of image grayscale being limited to a small range and improving the visual effect of the image. This paper uses scale_image operator performs grayscale stretching transformation, with the parameter Mult being the multiplied coefficient and add being the offset. The calculation formula is as follows:

$$Mult = \frac{255}{GMax - GMin}, Add = -Mult * GMin \tag{24}$$

Where, $GMax$ represents the maximum grayscale value of the digital image, and $GMin$ represents the minimum grayscale value.

Firstly, perform a grayscale transformation on the original collected image to make the overall image clearer and more delicate. Secondly, use the emphasize operator for contrast enhancement to obtain high-quality barcode images for easy subsequent reading and recognition. The enhancement effect is shown in Fig. 6.

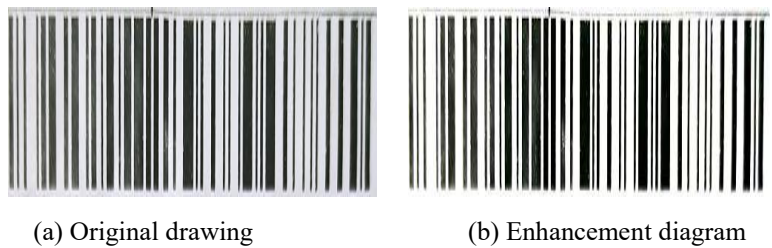

(a) Original drawing      (b) Enhancement diagram

**Fig. 6. Image** enhancement effect.

500 control boards with barcode pasting were tested, and all recognition results were correct as shown in Table 2.

Table 2. Barcode recognition test

| Serial Number | Barcode characters | recognition result | accuracy |
|---|---|---|---|
| 1 | 3007938949210 | 3007938949210 | 100% |
| 2 | 773169579357455 | 773169579357455 | 100% |
| 3 | 800129365280 | 800129365280 | 100% |
| 4 | 800129362045 | 800129362045 | 100% |
| 5 | 800129355506 | 800129355506 | 100% |
| ⋮ | ⋮ | ⋮ | ⋮ |
| 500 | 8167454731616 | 8167454731616 | 100% |

## 3.7. Algorithm simulation and analysis

This paper developed a motor control board defect detection system based on HALCON software and MVtec Deep Learning Tool, and conducted simulation verification and analysis using publicly available image dataset (https://github.com/tangsanli5201/DeepPCB)[43]. The comparative verification results of different defects are shown in Fig. 7.

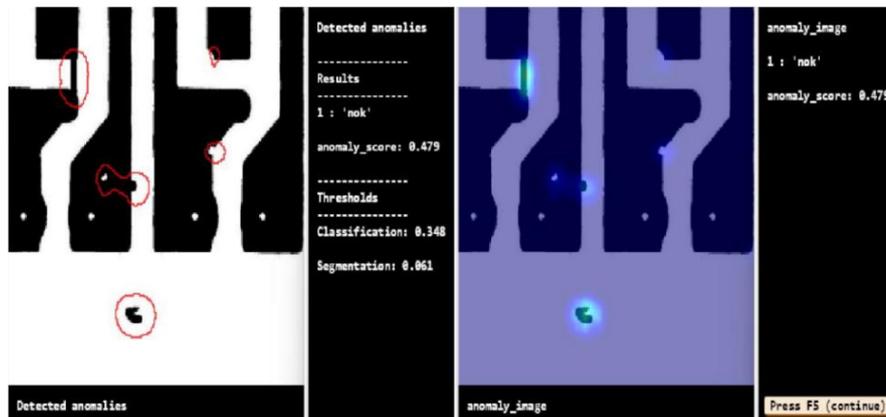

(a) Defect type (open circuit, stray, short circuit)

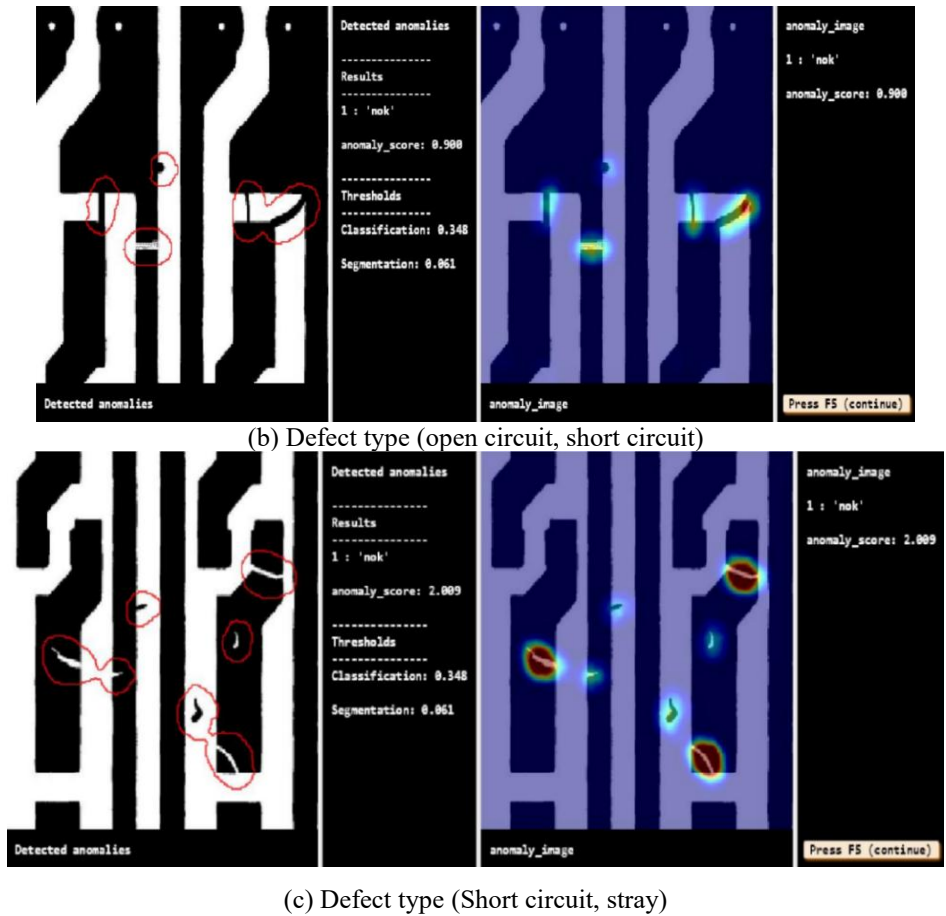

(b) Defect type (open circuit, short circuit)

(c) Defect type (Short circuit, stray)

**Fig. 7.** Defect identification results.

This paper successfully processed 800 sample images with an accuracy rate of over 99%, demonstrating the precise recognition ability of defect detection methods for defect types. The corresponding products selected in this experiment are widely representative, and the experimental results and time tolerance test show that the algorithm runs completely and the experimental results are reliable, which can be applied to actual production.

## 4. Results and discussion

### 4.1. Experimental procedure

According to the workflow of the intelligent visual inspection system, the defect detection and processing flow is as follows:

(1) Start the machine vision detection system and open a 512×512 pixel black detection window.

(2) The transmission module transports the control board to the position to be detected, and the camera promptly collects and acquires image information.

(3) Display the original image, and convert the color image to a grayscale image.

(4) Define and initialize variables ROI_0 and ROI_1 for the region of interest, merge them into RegionUnion, and display the region of interest in the window.

(5) Define and initialize the variable RGB_Image, decompose it into three channels: R, G and B.

(6) Convert RGB images into HSV space.

(7) Extract the color features of region ROI_0, including hue, saturation and brightness, and display them in the window.

(8) Extract the color features of region ROI_1, including hue, saturation, and brightness, and display them in the window.

(9) Determine whether the color difference of the control board on the assembly line is qualified, and set a threshold to determine whether the brightness characteristics of regions ROI0 and ROI1 are greater than 150. If it is greater than 150, it is considered a qualified product, otherwise it is considered a defective product and the judgment result is displayed in the window.

(10) Defective products with color differences are directly removed by the robotic arm, while control boards with qualified color differences enter the process of identifying surface defects.

Conduct experimental verification on the typical motor control board, extract image features of areas of interest, and determine whether the area is a qualified or defective product based on the features. The color difference recognition processing result of the motor control board is shown in Fig. 8.

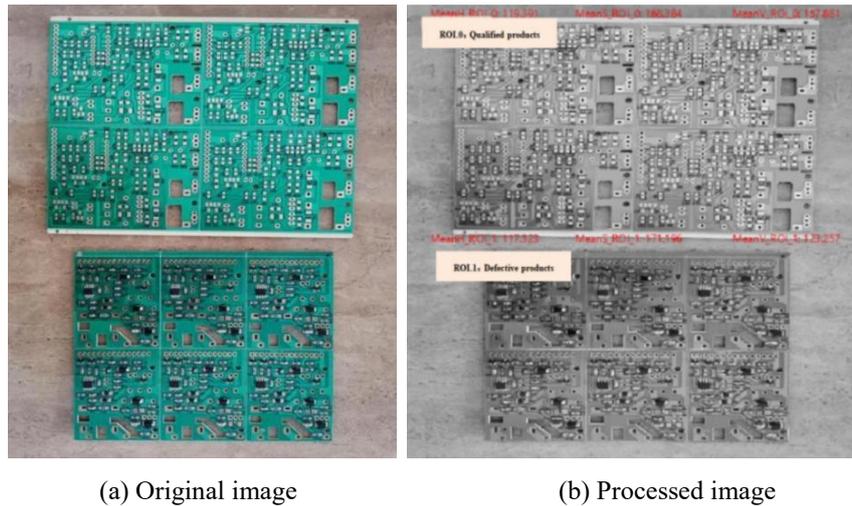

(a) Original image  (b) Processed image
**Fig. 8.** Comparison results of the motor control board color difference recognition.

## 4.2. Comparative analysis of experimental results

(1) Pre-Experiment

This paper divides 1000 pieces motor control boards into standard group, defect group, and color difference group, and uses correct images as reference calibration boards. Standard group: 600 pieces that meet the requirements. Defect group: 200 pieces with dimensions, positions, barcode, and other characteristics that are not qualified. Color difference group: all other features are qualified, but there is significant color difference, a control group of 200 pieces is set. Considering the successful acquisition of images from the establishment of the experimental platform, it is necessary to conduct pre-experiments and gradually adjust the angle and height of the camera to ensure the stability of the experimental platform and the correctness of the experimental results.

Firstly, this paper grouped the experimental subjects, labeled 1000 products with numbers 1-1000, and then divided them into 4 groups as shown in Table 3.

Table 3. Experimental groups

| Number of groups | Standard group | Defect group | Color difference group | Total |
|---|---|---|---|---|
| 1 | 150 | 50 | 50 | 250 |
| 2 | 150 | 50 | 50 | 250 |
| 3 | 150 | 50 | 50 | 250 |
| 4 | 150 | 50 | 50 | 250 |

Secondly, conduct a pre-experiment. Take any group of samples for stability testing, test whether the experimental platform is stable, conduct 100 experiments.

The summary of the pre experimental results is shown in Table 4.

Table 4. Summary table of pre-experimental results

| Number of experiments | Standard group | Correct identification | Defect group | Correct identification | Color difference group | Correct identification | Accuracy |
|---|---|---|---|---|---|---|---|
| 100 | 15000 | 14992 | 5000 | 4988 | 5000 | 4981 | 99.62% |

The accuracy of the pre experiment reached 99.62%, which met the experimental standard. After Log-rank, it was found that the number of incorrect identification occurred after 200 times, which met the production requirements.

(2) Group experiment

Conduct experiments in batches under the same conditions. The experiment was divided into two parts, the first part was a test which was conducted with four groups, and the second part was a long-term test which tested the standard test for 8h.

The results of the first experiments are shown in Table 5.

Table 5. Results of the first experiment

| Number of groups | Standard group | Correct identification | Defect group | Correct identification | Color difference group | Correct identification | Accuracy |
|---|---|---|---|---|---|---|---|
| 1 | 150 | 150 | 50 | 50 | 50 | 50 | 100% |
| 2 | 150 | 150 | 50 | 50 | 50 | 50 | 100% |
| 3 | 150 | 150 | 50 | 50 | 50 | 50 | 100% |
| 4 | 150 | 150 | 50 | 50 | 50 | 50 | 100% |

The first experiment used a small batch of products to test the stability of the algorithm and the reliability of the experimental results. The detection accuracy indicates that the second experiment can be conducted.

The results of the second experiments are shown in Table 6.

Table 6. Results of the second experiment

| Number of groups | Standard group | Correct identification | Defect group | Correct identification | Color difference group | Correct identification | Accuracy |
|---|---|---|---|---|---|---|---|
| 1 | 1500 | 1498 | 500 | 500 | 500 | 500 | 99.87% |
| 2 | 1500 | 1496 | 500 | 500 | 500 | 500 | 99.73% |
| 3 | 1500 | 1499 | 500 | 500 | 500 | 500 | 99.93% |
| 4 | 1500 | 1495 | 500 | 500 | 500 | 500 | 99.67% |

The corresponding products selected in the second experiment are widely representative, and the experimental results and time tolerance test show that the algorithm runs completely and the experimental results are reliable, which can be applied to actual production. In the future, we will apply the algorithm to defect detection of other products based on engineering practice.

(3) Visualization display of experimental results

The intelligent visual inspection system can quickly and effectively handle control board defects. By extracting the features of specific regions of interest in the image and making judgments based on pre-set thresholds, it is possible to quickly and accurately determine whether the control board is qualified, and to remove unqualified products through PLC controlled robotic arms. In the current detection process, it is easy to detect different types of control boards by simply modifying the corresponding regions of interest. This method provides an effective quality control tool for control board manufacturers. Fig. 9 shows the detection results of a typical control board.

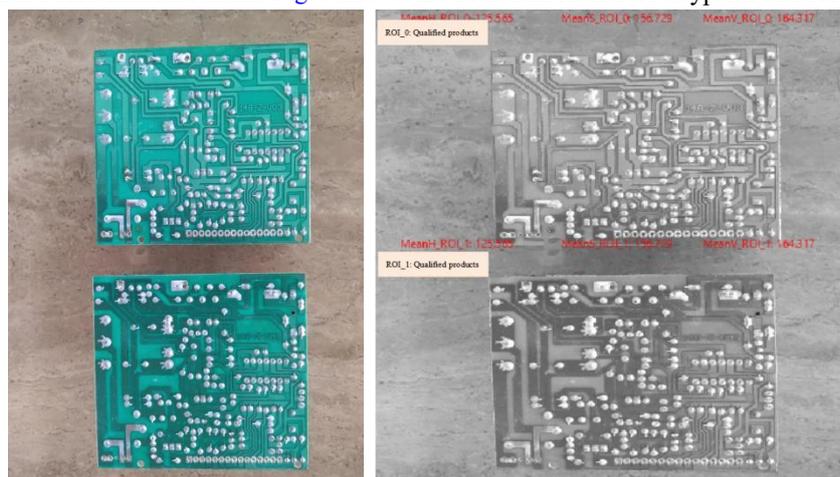

(a) Original image            (b) Processed image
**Fig. 9.** Visual inspection results of control board.

## 4.3. Discussion

This paper judges the two regions of interest based on preset thresholds to determine whether ROI_0 and ROI_1 are qualified products. For each region of interest, it checks whether the average characteristic value of the region is greater than a threshold. If it is greater than the threshold, the area is considered a qualified product and corresponding information is displayed in the window; Otherwise, consider the area as defective and display the corresponding information in the window. The process of selecting regions of interest involves calculating the parameters of rectangles and rotated rectangles in computational geometry, such as center point coordinates, length and width, rotation angle, etc. Therefore, the region of interest selection model can be regarded as an application of computational geometry models. Through the application of this model, the program can quickly locate the area of interest and perform subsequent processing. The threshold segmentation model in this article can segment an image into different regions and distinguish between qualified and defective products by thresholding the specific features of the regions of interest.

The color difference detection method can improve the quality control efficiency and accuracy of the motor control board. This method utilizes image processing techniques to first extract color features of specific regions of interest in the circuit board, and then judge the color based on a preset threshold to quickly and accurately determine whether the motor control board is qualified. In practical engineering applications, users only need to modify the corresponding areas of interest to easily detect different types of motor control boards, greatly improving the flexibility and scalability of detection.

Compared with traditional defect detection methods for motor control boards, this paper achieves comprehensive and multi angle detection of motor control boards through the use of image processing technology and color feature extraction algorithms, with high accuracy and strong robustness, and has broad application prospects and market potential.

## 5. Conclusions

The paper studied a visual image processing model and defect recognition method for motor control board defect detection. Compared to traditional motor control board detection methods, the proposed image processing model and feature extraction algorithm enable defect detection from all directions and angles, with high accuracy and strong robustness. The following research conclusions were obtained:

(1) A digital image processing model was established, and the key role of noise suppression in improving detection accuracy was analyzed. The experimental results showed that the digital image processing model was suitable for motor control board detection.

(2) The ideas for identifying and processing motor control board defects were discussed, specific methods for identifying various types of motor control board defects were proposed, image features from regions of interest were extracted, and qualified or defective products based on feature thresholds could be determined.

(3) The defect detection model based on machine vision has demonstrated high accuracy, effectively identifying defect types with an accuracy rate exceeding 99%. This means that the model can quickly process a large number of digital images of motor control boards in practical engineering applications, thereby achieving efficient defect detection.

The defect detection technology for motor control boards based on image processing is not only applicable to the field of electronic manufacturing, but can also be widely applied to defect detection of integrated circuit boards, performance testing and assembly of integrated electronic systems. It also has positive significance for the appearance inspection, defect diagnosis and reliability evaluation of electronic components. With the continuous advancement of machine vision technology and the expansion of its application fields, this technology will undoubtedly provide more advanced detection capabilities, wider application scenarios, and deeper socio-economic significance for manufacturing circuit boards or control boards. The future research of the paper will introduce deep learning and edge computing technologies to enhance the robustness of the model, improve the detection accuracy of the model for minor defects, so that the machine vision system can better process massive image data, and achieve faster and more accurate image analysis and processing. Integrating cloud computing and cloud resource utilization technologies to achieve more efficient and stable product quality control.

**CRediT authorship contribution statement**

**Jingde Huang:** Writing–original draft, Writing–review & editing, Methodology, Resources, Project administration, Funding acquisition. **Zhangyu Huang:** Writing – review & editing, Validation, Supervision. **Chenyu Li**: Software, Formal analysis, Data curation, Conceptualization. **Jiantong Liu: Resources**, Investigation.

**Declaration of competing interest**

The authors declare that they have no known competing financial interests or personal relationships that could have appeared to influence the work reported in this paper.

**Acknowledgements**

This research was funded by Key Foundation of universities in Guangdong Province, China, grant numbers 2024ZDZX3003 and 2024ZDJS135; Zhuhai Basic and Applied Basic Research Project, grant number 2320004002337.

**Supplementary data**

Supplementary data to this article can be found online at https://github.com/tangsanli5201/DeepPCB.